\documentclass{article}

\usepackage[preprint]{neurips_2026}

\usepackage[utf8]{inputenc} 
\usepackage[T1]{fontenc}    
\usepackage{booktabs}       
\usepackage{amsmath}        
\usepackage{amssymb}        
\usepackage{amsfonts}       
\usepackage{amsthm}         
\usepackage{mathtools}      
\usepackage{bm}             
\usepackage{algorithm}      
\usepackage{algpseudocode}  
\usepackage{nicefrac}       
\usepackage{microtype}      
\usepackage{xcolor}         
\usepackage{graphicx}       
\usepackage{float}          
\usepackage{placeins}       
\usepackage{url}            
\usepackage{hyperref}       

\graphicspath{{figures/}{tex/figures/}{../presentations/}{presentations/}}

\newtheorem{theorem}{Theorem}
\newtheorem{proposition}{Proposition}

\theoremstyle{remark}

\newcommand{\R}{\mathbb{R}}
\newcommand{\norm}[1]{\left\lVert #1 \right\rVert}
\newcommand{\softmax}{\operatorname{softmax}}
\newcommand{\TopK}{\operatorname{TopK}}
\newcommand{\TopM}{\operatorname{TopM}}

\title{SDG-MoE: Signed Debate Graph Mixture-of-Experts}

\author{Stepan Kulibaba \\
  Innopolis University, Innopolis; \\
  Sirius University, Sirius \\
  \texttt{kulibabast@gmail.com} \\
  \And
  Kirill Labzin \\
  Sirius University, Sirius \\
  \texttt{labzin.kr@gmail.com} \\
  \And
  Artem Dzhalilov \\
  Innopolis University, Innopolis; \\
  Sirius University, Sirius \\
  \texttt{artem.dzhalilov@gmail.com} \\
  \And
  Roman Pakhomov \\
  Innopolis University, Innopolis \\
  \texttt{r2087007@gmail.com} \\
  \And
  Oleg Svidchenko \\
  HSE University \\
  \texttt{argentumwalker@gmail.com} \\
  \And
  Alexander Gasnikov \\
  Innopolis University, Innopolis \\
  \texttt{gasnikov@yandex.ru} \\
  \And
  Aleksei Shpilman \\
  HSE University  \\
  \texttt{alexey@shpilman.com} \\
}

\begin{document}

\maketitle

\begin{abstract}
Sparse MoE models achieve a good balance between capacity and compute by routing each token to a small subset of experts. However, in most MoE architectures, once a token is routed, the selected experts process it independently and their outputs are combined via a weighted sum. This leaves open whether enabling communication among them could improve performance. While prior work has raised this question, direct interaction among the active routed experts remains underexplored. In this paper, we propose SDG-MoE (Signed Debate Graph Mixture-of-Experts), a novel architecture that adds a lightweight, iterative deliberation step before final aggregation. SDG-MoE introduces three components: (i) two learned interaction matrices over the active experts, a support graph $A^+$ and a critique graph $A^-$, capturing reinforcing and corrective influences; (ii) a signed message-passing step that updates expert representations before aggregation; and (iii) a disagreement-gated Friedkin--Johnsen-style anchoring that controls deliberation strength while preventing expert drift. Together, these enable a structured deliberation process where interaction strength scales with disagreement and specialization is preserved. We also provide a theoretical analysis establishing stability conditions on expert states and showing that deliberation adds only low-order overhead over the active set. In controlled three-seed pretraining experiments, SDG-MoE improves validation perplexity over both an unsigned graph communication baseline and vanilla MoE, outperforming the strongest baseline by 19.8\%, and gives the best external perplexity on WikiText-103, C4, and Paloma among the compared systems.
\end{abstract}

\section{Introduction}
Large language models benefit from scaling \cite{scalinglaws}, but dense Transformer architectures come with high computational cost since all parameters are activated for each token. Sparse Mixture-of-Experts (MoE) layers address this by activating only a small subset of experts, improving the capacity–compute trade-off.

\begin{figure}[t]
    \centering
    \includegraphics[width=\linewidth]{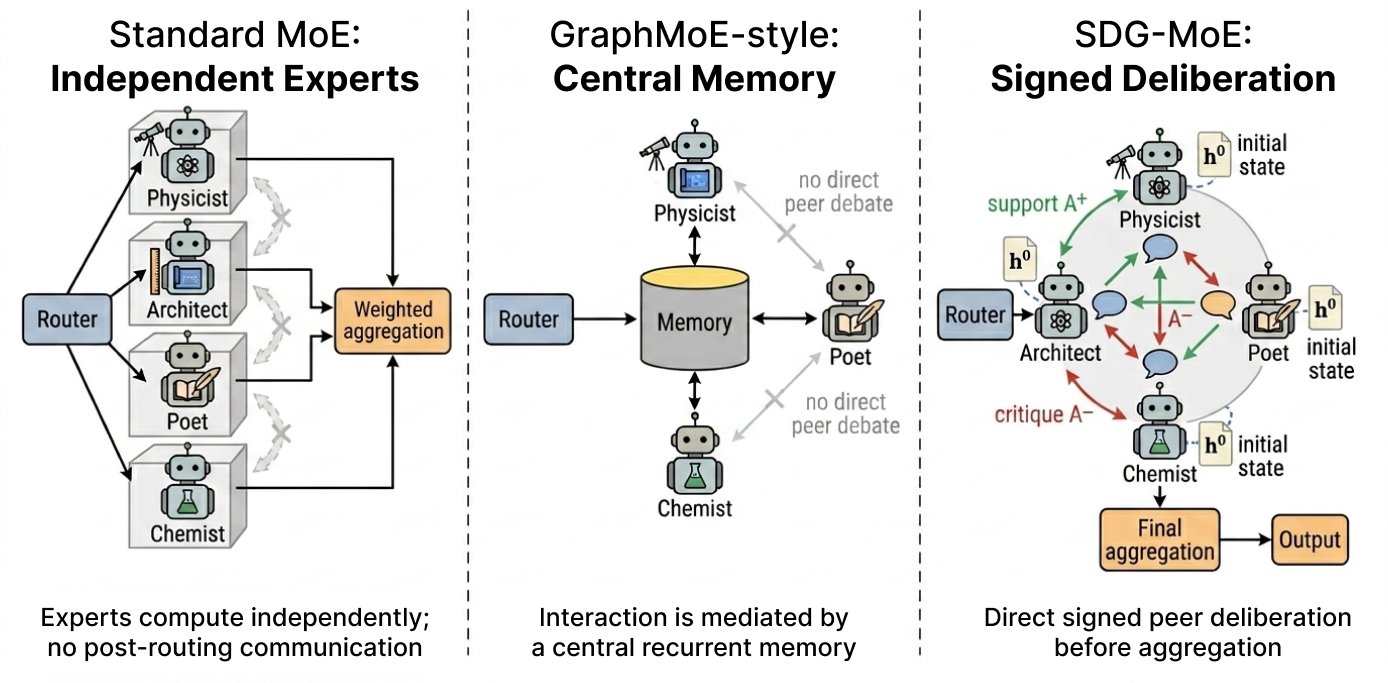}
    \caption{Conceptual comparison of post-routing expert interaction. Standard MoE combines expert outputs that are computed independently. GRAPHMOE-style methods allow experts to exchange information, but usually through a central memory or virtual node. SDG-MoE instead views the active experts as a signed social system, where routed experts can directly support or critique each other before the final aggregation.}
    \label{fig:overview-pipelines}
\end{figure}

In standard Top-$K$ MoE, selected experts process tokens independently and their outputs are combined using a weighted average. This simple aggregation does not simulate interactions between experts, which may be informative for the final representation. Prior work has studied richer ways to combine expert outputs, including attention-based fusion and auxiliary communication mechanisms~\cite{hypermoe,mode,GRAPHMOE}. However, these approaches typically do not model the selected experts as a fully interacting system at inference time, where each expert can iteratively influence others. Figure~\ref{fig:overview-pipelines} shows this progression from independent expert voting, to centrally mediated interaction, to the signed peer deliberation proposed in our work.

We suggest a different view of the selected experts: not as independent predictors but as a small interacting group. Motivated by the expert committees and social deliberation, we hypothesize that routed experts can form stronger representations if they are allowed to support, critique, and revise one another before aggregation. In this view, expert states are not fixed after routing. Instead, they are refined through structured communication, which helps capture richer patterns of agreement, disagreement, and specialization before the final aggregation.

We propose \emph{Signed Debate Graph Mixture-of-Experts} (SDG-MoE), a sparse MoE layer that augments Top-$K$ routing with a lightweight communication phase. SDG-MoE has three key components: \textbf{signed expert graphs}, \textbf{anchored deliberation}, and \textbf{private/shared expert states}. For each routed token, the active experts construct two learned interaction matrices, a support graph $A^+$ and a criticism graph $A^-$, and update their shared states by signed message passing before aggregation. The update follows a Friedkin--Johnsen-type \cite{friedkin1990social} dynamics: experts can influence one another, but each expert remains anchored to its initial representation. A disagreement-based gate decides how strongly experts should communicate. At the same time, private states skip this social update, helping preserve expert specialization while the shared states are refined through deliberation.

Our primary contributions:
\begin{itemize}
\item We introduce, to our knowledge, the first sparse MoE layer that enables direct signed communication among the active routed experts before aggregation. In controlled three-seed pretraining runs, SDG-MoE improves validation perplexity over both a matched unsigned graph baseline and vanilla MoE, outperforming the strongest baseline by 19.8\%, and yields the strongest external perplexity among the compared systems.
\item We provide a theoretical analysis of signed expert deliberation, establishing stability conditions under which expert states remain bounded and showing that the additional communication cost remains low-order in the active set size.
\item We connect classical multi-agent consensus and social opinion dynamics with modern MoE architectures, yielding a compact design based on support/critique graphs, Friedkin--Johnsen-style anchoring, and private/shared expert states.
\end{itemize}

\section{Related Work}

\subsection{Sparse Mixture-of-Experts}
Sparse Mixture-of-Experts has become a central approach for scaling language models under a fixed computational budget. Building on classical conditional computation, sparse MoE layers replace dense feed-forward computation with a pool of experts, only a small subset of which is activated for each token. Large-scale architectures such as Sparsely-Gated MoE \cite{sparselygatedmoe}, GShard \cite{gshard}, Switch Transformer \cite{switchtransformer}, GLaM \cite{glam}, ST-MoE \cite{stmoe}, Mixtral \cite{mixtral}, and DeepSeekMoE \citet{deepseekmoe} show that this design can substantially increase model capacity while keeping per-token computation manageable. Most work in this direction focuses on routing, load balancing, expert utilization, training stability, and efficient distributed execution \cite{Cai_2025}, establishing sparse MoE as a practical scaling mechanism for modern language models.

\subsection{Classical Opinion Dynamics}
Classical opinion dynamics studies how agents update beliefs through social influence. In DeGroot \cite{degroot1974reaching} models, agents repeatedly average neighbors’ opinions, leading to consensus. Friedkin–Johnsen \cite{friedkin1990social} extends this by combining social influence with persistent private beliefs, making it especially suitable for MoE communication: selected experts can exchange information while retaining their specialized representations. While bounded-confidence and signed-network models capture selective influence and disagreement, they are typically defined over fixed social graphs and low-dimensional opinions; our setting adapts the Friedkin–Johnsen intuition to learned expert states after routing.

\subsection{Expert Interaction and Coordination in MoE}

Recent work has begun to investigate interaction and coordination in sparse MoE by transferring information across experts. HyperMoE \cite{hypermoe} incorporates information from non-selected experts through a hypernetwork, although this information is compressed into an auxiliary signal rather than exchanged directly among selected experts. MoDE \cite{mode} promotes agreement through mutual distillation, but the interaction is embedded in the training objective and does not alter inference-time expert computation. SymphonySMoE \cite{nguyennhat2025modelingexpertinteractionssparse} uses a social graph between experts to guide co-selection via smoothed routing, but without enabling communication between the selected experts. GRAPHMOE \cite{GRAPHMOE} introduces inference-time coordination through a centralized shared memory node that collects and redistributes information across recurrent reasoning rounds, rather than through direct expert-to-expert communication. Chain-of-Experts (CoE) \cite{chainofexperts} avoids such centralized coordination by passing intermediate representations through a sequential chain of routed experts, but communication is organized as step-wise refinement rather than joint deliberation among a fixed active expert set. These approaches broaden the design space of MoE beyond isolated expert computation, yet explicit post-routing interaction among selected experts remains comparatively underexplored, particularly when experts must exchange conflicting signals and move toward a shared consensus before aggregation.

\section{SDG-MoE}
\label{sec:architecture}

\begin{figure}[t]
    \centering
    \includegraphics[width=\linewidth]{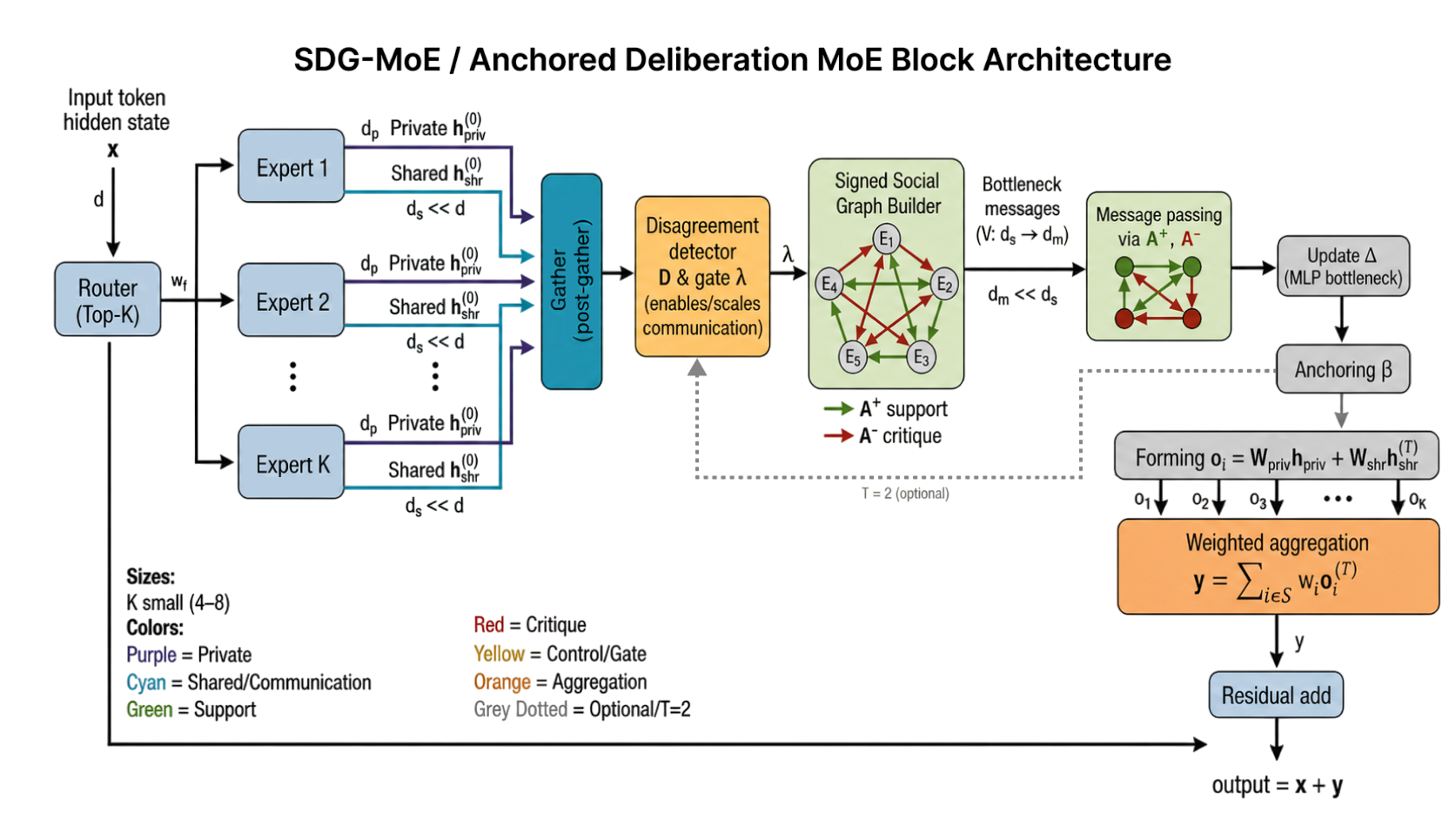}
    \caption{SDG-MoE augments Top-$K$ routing with a compact signed deliberation module over the active experts. Only the shared expert states communicate, while private states bypass the graph and preserve specialization.}
    \label{fig:sdg_moe_block}
\end{figure}

SDG-MoE extends a standard sparse MoE feed-forward block by adding a deliberation phase between routing and aggregation. It keeps the usual Top-$K$ router and changes only what happens between expert evaluation and aggregation. For a token state $x\in\R^d$, the router produces probabilities
\begin{equation}
    \pi=\softmax(W_r x), \qquad S=\TopK(\pi,K), \qquad
    w_i=\frac{\pi_i}{\sum_{j\in S}\pi_j}, \quad i\in S .
    \label{eq:sdg-routing}
\end{equation}
A vanilla MoE would return $\sum_{i\in S}w_iE_i(x)$. SDG-MoE instead treats the selected experts as a small deliberative group and lets them exchange low-dimensional signed messages before the final consensus. Figure~\ref{fig:sdg_moe_block} summarizes this block-level computation.

\subsection{Routed Experts with Private and Shared States}

The first step is intentionally close to a conventional MoE layer. Once the router has selected $S$, each active expert computes a hidden representation and exposes two outputs:
\begin{align}
    r_i &= \phi(W^{\mathrm{up}}_i x+b^{\mathrm{up}}_i),\\
    h^{(0)}_{i,\mathrm{priv}} &= W^{\mathrm{priv}}_i r_i+b^{\mathrm{priv}}_i,
    \qquad
    h^{(0)}_{i,\mathrm{shr}} = W^{\mathrm{shr}}_i r_i+b^{\mathrm{shr}}_i .
    \label{eq:expert-split}
\end{align}
where $W^{\mathrm{priv}}_i \in \mathbb{R}^{(d-d_s)\times d_r}$ and
$W^{\mathrm{shr}}_i \in \mathbb{R}^{d_s\times d_r}$, with $d_s \ll d$ to preserve a large private subspace for specialization.
By design, only $h_{i,\mathrm{shr}}$ participates in message passing,
while $h_{i,\mathrm{priv}}$ is carried unchanged to the output.

\subsection{Signed Debate Graph}
At deliberation round $t$, SDG-MoE builds two directed graphs on the active set $S$: a support graph $A^{+,(t)}$ and a criticism graph $A^{-,(t)}$. Let
\begin{equation}
    z_i^{(t)} = [\,\operatorname{LayerNorm}(h^{(t)}_{i,\mathrm{shr}}); e_i\,],
\end{equation}
where $e_i$ is a learned expert identity embedding. Separate query/key projections into a $d_g$-dimensional graph space define the two edge types:
\begin{align}
    A_{ij}^{+,(t)}
    &= \softmax_j\!\left(
        \frac{\langle W_Q^+z_i^{(t)}, W_K^+z_j^{(t)}\rangle}{\sqrt{d_g}} + B^+_{ij}
    \right),\\
    \widetilde A_{ij}^{-,(t)}
    &= \softmax_j\!\left(
        \frac{\langle W_Q^-z_i^{(t)}, W_K^-z_j^{(t)}\rangle}{\sqrt{d_g}} + B^-_{ij}
    \right).
    \label{eq:signed-graphs}
\end{align}
The masks $B^+$ and $B^-$ control self-loops: by default we allow support self-loops and remove criticism self-loops. To keep criticism targeted rather than diffuse, each row of $\widetilde A^-$ is sparsified to its strongest $m_-$ entries and renormalized:
\begin{equation}
    A_{i\cdot}^{-,(t)}
    =
    \operatorname{Normalize}\!\left(\TopM_{m_-}(\widetilde A_{i\cdot}^{-,(t)})\right).
    \label{eq:negative-topm}
\end{equation}
Here $\TopM_{m_-}$ keeps the $m_-$ largest entries in row $i$ and sets all
remaining entries to zero. The operator $\operatorname{Normalize}$ then divides
the surviving weights by their row sum, with a small numerical $\varepsilon$ in
the denominator. Thus the negative graph remains row-stochastic after
sparsification, while each expert criticizes only its $m_-$ strongest negative
neighbors.

\subsection{Gated Deliberation and Anchoring}

Communication should matter most when selected experts actually disagree. SDG-MoE therefore computes a bounded disagreement score from a low-dimensional projection via a learned matrix $P_D$:
\begin{equation}
    \widehat h_i^{(t)}=
    \frac{P_D h_{i,\mathrm{shr}}^{(t)}}{\norm{P_D h_{i,\mathrm{shr}}^{(t)}}_2+\varepsilon},
    \qquad
    D^{(t)} =
    \sqrt{\frac{1}{K(K-1)}
    \sum_{i\ne j}\frac{1-\langle \widehat h_i^{(t)},\widehat h_j^{(t)}\rangle}{2}} .
    \label{eq:disagreement}
\end{equation}
With the default gate, the social step size is
\begin{equation}
    \lambda^{(t)}
    =
    \lambda_{\min}
    +(1-\lambda_{\min})\tanh\!\left(a[D^{(t)}-\delta]_+\right),
    \label{eq:social-gate}
\end{equation}
so deliberation is suppressed below the threshold $\delta$ and grows smoothly as the active experts become more polarized. An optional expert-specific confidence gate $g_i(x)\in[0,1]$ further scales the update for expert $i$ on the current token.

Shared states are projected into message vectors $u_j^{(t)}=Vh^{(t)}_{j,\mathrm{shr}}$. Each expert receives support and criticism messages
\begin{equation}
    m_i^{+,(t)}=\sum_{j\in S}A_{ij}^{+,(t)}u_j^{(t)},\qquad
    m_i^{-,(t)}=\sum_{j\in S}A_{ij}^{-,(t)}u_j^{(t)} .
\end{equation}
The update MLP sees the current shared state, the support message, and the signed contrast:
\begin{equation}
    \Delta_i^{(t)}
    =
    U_2\phi\!\left(
    U_1[\,h_{i,\mathrm{shr}}^{(t)};m_i^{+,(t)};m_i^{+,(t)}-\gamma m_i^{-,(t)}\,]+b_1
    \right)+b_2 .
    \label{eq:sdg-update-delta}
\end{equation}
Although the update MLP could in principle learn an equivalent linear reparameterization from $(m_i^+,m_i^-)$, the explicit contrast input biases the module toward treating the negative channel as counter-evidence rather than as a second unsigned message. The shared state is then updated with a gated residual step followed by anchoring to the initial shared opinion:
\begin{align}
    \bar h_{i,\mathrm{shr}}^{(t+1)}
    &= h_{i,\mathrm{shr}}^{(t)}+\alpha\lambda^{(t)}g_i(x)\Delta_i^{(t)},\\
    h_{i,\mathrm{shr}}^{(t+1)}
    &= \beta h_{i,\mathrm{shr}}^{(0)}
    +(1-\beta)\bar h_{i,\mathrm{shr}}^{(t+1)} .
    \label{eq:anchored-update}
\end{align}
The coefficient $\beta$ plays the role of a Friedkin--Johnsen stubbornness parameter: larger values make an expert rely more on its initial shared state, while smaller values allow stronger influence from other active experts. We use a small number of rounds $T$ and, by default, rebuild the graph after each round so that the social structure follows the evolving shared states. Algorithm~\ref{alg:sdg-round} in the appendix gives the complete one-round update.

\subsection{Output Aggregation}

After deliberation, the $i$-th active expert output is
\begin{equation}
    o_i^{(T)} =
    W^{\mathrm{out,priv}}_i h^{(0)}_{i,\mathrm{priv}}
    + W^{\mathrm{out,shr}}_i h^{(T)}_{i,\mathrm{shr}} .
    \label{eq:expert-output}
\end{equation}
These output projections are expert-specific, as in a standard MoE down-projection. We aggregate the resulting outputs with the router weights,
\begin{equation}
    y=\sum_{i\in S}w_io_i^{(T)}.
\end{equation}
The Transformer block then applies the usual residual update $x\leftarrow x+y$.

\section{Theoretical Analysis of Deliberation Dynamics}
\label{sec:theory}

We analyze one SDG-MoE layer after the router has selected an active set of
$K$ experts. Let $H^{(t)}\in\mathbb{R}^{K\times d_s}$ collect their shared
states. One deliberation round can be written as
\begin{equation}
    H^{(t+1)}=
    \beta H^{(0)}+(1-\beta)\!\bigl[
    H^{(t)}+\alpha\lambda(D(H^{(t)}))G(x)\Delta_x(H^{(t)})
    \bigr],
    \label{eq:abstract-sdg}
\end{equation}
where $\Delta_x$ denotes the signed message-passing update, $G(x)=\operatorname{diag}(g_i(x))_{i\in S}$ is the
expert-specific confidence gate, $\lambda(D)$ is the disagreement gate, and $\beta$ anchors
experts to their initial routed states. Thus the theory has three goals: show
that deliberation is a low-order active-set correction, that anchoring controls
the recurrent update, and that the hyperparameters have interpretable roles.
All proofs are given in Appendix~\ref{app:theory}.

\paragraph{Local assumptions.}
We use only finite-round, local regularity assumptions. On the states visited by
Eq.~\eqref{eq:abstract-sdg}, assume
\begin{equation}
\begin{aligned}
    \|\Delta_x(H)\|_F &\le B_\Delta,\\
    \|\Delta_x(H)-\Delta_x(\widetilde H)\|_F
    &\le L_\Delta\|H-\widetilde H\|_F,
\end{aligned}
\label{eq:local-assumptions}
\end{equation}
and let $D$ and $\lambda$ be Lipschitz with constants $L_D$ and $L_\lambda$.
Assume also $0\le g_i(x)\le g_{\max}\le 1$ for every active expert and
$\max_i\|W^{\mathrm{out,shr}}_i\|_2\le B_W$. These conditions are local to the bounded
trajectory induced by LayerNorm, normalized cosine disagreement, bounded gates, and the small fixed number of deliberation rounds. This interpretation is discussed in Appendix~\ref{app:assumptions}.

\begin{proposition}[Low-order active-set overhead]
\label{prop:cost}
Let $N$ denote the total number of experts in the layer and
$b=\max\{d_s,d_g,d_m,d_{\mathrm{dis}},d_{\mathrm{upd}},d_e\}$ collect the
SDG bottleneck widths. One communication round costs
$O(Kb^2+K^2b)$ per token. With $T$ rounds and parameters shared across rounds,
\begin{equation}
    F_{\mathrm{SDG}}(T)=O\bigl(T(Kb^2+K^2b)\bigr),
    \qquad
    P_{\mathrm{SDG}}=O(b^2+Nd_e).
\end{equation}
Relative to a Top-$K$ expert FFN cost $F_{\mathrm{MoE}}=\Theta(Kdd_{\mathrm{ff}})$,
\begin{equation}
    \frac{F_{\mathrm{SDG}}(T)}{F_{\mathrm{MoE}}}
    =O\!\left(T\frac{b^2+Kb}{dd_{\mathrm{ff}}}\right).
    \label{eq:relative-cost}
\end{equation}
\end{proposition}
Hence, for small $K,T$ and $b\ll d,d_{\mathrm{ff}}$, SDG adds a compact
communication correction rather than another large expert layer.

\begin{theorem}[Controlled and stable deliberation]
\label{thm:stability}
For Eq.~\eqref{eq:abstract-sdg} with $0<\beta<1$ and any finite $T$,
\begin{equation}
    \begin{aligned}
    \|H^{(T)}-H^{(0)}\|_F
    &\le
    \frac{(1-\beta)\alpha g_{\max}B_\Delta}{\beta}
    \bigl(1-(1-\beta)^T\bigr),\\
    \|y_T-y_0\|_2
    &\le
    B_W\frac{(1-\beta)\alpha g_{\max}B_\Delta}{\beta}
    \bigl(1-(1-\beta)^T\bigr).
    \end{aligned}
    \label{eq:bounded-hidden}
\end{equation}
Moreover, with $C=L_\Delta+B_\Delta L_\lambda L_D$, the one-round map is a contraction whenever
\begin{equation}
    q=(1-\beta)(1+\alpha g_{\max}C)<1
    \quad\Longleftrightarrow\quad
    \alpha g_{\max}C < \frac{\beta}{1-\beta}.
    \label{eq:compact-contraction}
\end{equation}
In this regime, there is a unique deliberation fixed point $H^\star$ and
$\|H^{(T)}-H^\star\|_F\le q^T\|H^{(0)}-H^\star\|_F$.
\end{theorem}

The theorem shows that SDG is a controlled perturbation of vanilla MoE: larger
$\alpha$ and $T$ permit stronger communication, while larger $\beta$ keeps the
model close to the routed expert opinions and can certify convergence. The same
bound explains why excessive rounds may be wasteful: cost grows linearly in $T$,
whereas the residual distance to the stable point decays geometrically as $q^T$.

\begin{proposition}[Signed critique and hyperparameter interpretation]
\label{prop:targeted-critique}
SDG-MoE exposes both support and signed contrast,
$(m_i^+,m_i^+-\gamma m_i^-)$ , so the update can distinguish confirming evidence from counter-evidence. In the centered outlier model of Appendix~\ref{app:signed-proof}, if an inlier assigns $A_{io}^-\ge\eta$ and $A_{io}^+\le\xi$ to an outlier direction $r$, then the signed contrast has negative projection on $r$ whenever
\[
    \gamma\eta>\xi
    \quad\text{and}\quad
    \|r\|>\frac{(1+\gamma)\epsilon}{\gamma\eta-\xi}.
\]
The disagreement threshold is calibrated by
\begin{equation}
    D^2=\tfrac12(1-\overline{\cos}),
    \qquad
    D>\delta \iff \overline{\cos}<1-2\delta^2.
\end{equation}
For the default gate
\[
    \lambda(D)=\lambda_{\min}+(1-\lambda_{\min})\tanh(a[D-\delta]_+),
\]
we have $L_\lambda\le (1-\lambda_{\min})a$. Therefore a conservative stability certificate is preserved when
\begin{equation}
    (1-\beta)\left[
    1+\alpha g_{\max}
    \bigl(L_\Delta+B_\Delta(1-\lambda_{\min})aL_D\bigr)
    \right]<1 .
    \label{eq:stable-gate-main}
\end{equation}
Thus larger $a$, $\alpha$, or $g_{\max}$ requires stronger anchoring $\beta$ for
the same stability margin.
\end{proposition}

\noindent
Together, these relations make the main deliberation controls interpretable:
$\delta$ and $a$ shape when communication activates, $\alpha$ sets its strength,
and $\beta$ keeps the dynamics controlled.

\section{Experiment}
\label{sec:experiments}
\subsection{Experimental Setup}
\paragraph{Backbone.}
We evaluate SDG-MoE in a controlled pretraining setting built around a Qwen-like
decoder-only Transformer \cite{qwen3}. Concretely, we start from the architectural recipe of
Qwen-style language models and scale the width, depth, attention heads, and feed-forward
dimensions to our experimental budget. Additional details on the scaled model
configuration and training protocol are given in
Appendix~\ref{app:model-config}.

\paragraph{Datasets and Metrics.}
The primary metric is validation perplexity on the SmolLM-Corpus \cite{smollm} mixture. To check whether the conclusions transfer beyond this cache, we additionally report external perplexity on WikiText-103 \cite{wikitext}, English C4 \cite{c4} validation, and a compact Paloma \cite{paloma} subset. WikiText-103 provides a standard academic language-modeling benchmark, C4 gives a broad web-text distribution, and Paloma tests fit across multiple English/code domains rather than a single monolithic corpus. We further report forward FLOPs/token, training and inference throughput, and internal deliberation diagnostics.

\paragraph{Baselines and controls.}
We organize comparisons into three groups. First, standard baselines test whether
the gains come from sparse routing or generic expert fusion: \emph{Dense
Transformer}, \emph{Vanilla MoE}, \emph{MLP Fusion} (an MLP over the router-weighted mean and across-expert standard deviation of expert outputs), and \emph{Set Attention} (multi-head self-attention over the active expert states).
Second, expert-interaction baselines compare SDG-MoE to prior mechanisms that
allow experts to coordinate at inference time, including \emph{GRAPHMOE} and
\emph{Chain-of-Experts}. Finally, SDG-specific controls isolate the role of signed
deliberation. \emph{Unsigned Graph} removes the support/critique distinction,
\emph{Dual-Unsigned} matches the two-channel capacity without signed semantics,
\emph{SDG fixed $\lambda g$} replaces the adaptive social coefficient by a
constant, and the eval-only SDG interventions test a trained checkpoint without
retraining: \emph{zero-neg} removes critique messages, \emph{swap-sign} exchanges
support and critique, and \emph{zero-pos} removes support messages.
Hardware and seed settings are listed in
Appendix~\ref{app:experiment-details}, together with a more detailed description
of the baselines and controls.

\subsection{Main Results}
Table~\ref{tab:main-results} gives the main controlled pretraining comparison. Against the matched sparse MoE backbone, SDG-MoE reduces validation perplexity from $63.14$ to $48.03$, with only about $14\%$ more forward FLOPs/token than Vanilla MoE. Inference throughput decreases from $45.0$k to $34.5$k tokens/s, so deliberation has a measurable cost, but the quality gain is large relative to the added active-set computation. The ranking also transfers: SDG-MoE is best on WikiText-103, C4, and Paloma, and the three-seed gains over Vanilla MoE and the matched unsigned graph are statistically significant under Welch tests.

Generic fusion and interaction baselines do not close the gap. MLP Fusion and Set Attention add capacity around the active set, while GRAPHMOE and Chain-of-Experts introduce heavier interaction, but all are weaker in this pretraining regime. This may reflect that those methods are better aligned with downstream refinement or centrally mediated reasoning, whereas SDG-MoE adapts the routed experts themselves: private states preserve specialization, and the signed shared channel lets experts correct one another before aggregation.

\begin{table}[t]
\centering
\caption{Main pretraining comparison. Perplexities are reported as mean $\pm$ std. over three runs, with lower values indicating better performance. External PPL columns use deterministic evaluation slices.}
\label{tab:main-results}
\resizebox{\linewidth}{!}{%
\begin{tabular}{lcccccccc}
\toprule
Method & Params & Fwd FLOPs/tok & Train Tok/s & Inf. Tok/s & Val PPL $\downarrow$ & WT103 PPL $\downarrow$ & C4 PPL $\downarrow$ & Paloma PPL $\downarrow$ \\
\midrule
\multicolumn{9}{l}{\textit{Standard baselines}} \\
Dense Transformer & 806.2M & 1.662G & \textbf{29.7k} & 156.3k & 59.96 $\pm$ 2.12 & 231.9 $\pm$ 8.4 & 154.8 $\pm$ 5.6 & 193.3 $\pm$ 7.0 \\
Vanilla MoE & 808.0M & 0.739G & 17.4k & 45.0k & 63.14 $\pm$ 0.56 & 258.3 $\pm$ 2.3 & 172.2 $\pm$ 1.5 & 215.2 $\pm$ 1.9 \\
MLP Fusion & 896.2M & 0.915G & 16.4k & 43.7k & 79.53 $\pm$ 5.36 & 248.6 $\pm$ 16.8 & 171.0 $\pm$ 11.6 & 209.8 $\pm$ 14.2 \\
Set Attention & 839.2M & 0.812G & 14.4k & 42.9k & 59.91 $\pm$ 1.03 & 236.4 $\pm$ 4.1 & 158.0 $\pm$ 2.8 & 197.2 $\pm$ 3.5 \\
\midrule
\multicolumn{9}{l}{\textit{Expert-interaction baselines}} \\
GRAPHMOE & 1.1B & 1.030G & 8.2k & 28.3k & 63.54 $\pm$ 3.81 & 251.2 $\pm$ 15.1 & 168.7 $\pm$ 10.1 & 210.0 $\pm$ 12.6 \\
Chain-of-Experts & 1.1B & 1.005G & 10.2k & 19.3k & 74.75 $\pm$ 2.43 & 295.6 $\pm$ 9.6 & 198.5 $\pm$ 6.5 & 247.1 $\pm$ 8.1 \\
\midrule
\multicolumn{9}{l}{\textit{SDG-MoE variants and controls}} \\
SDG fixed $\lambda g$ & 840.1M & 0.842G & 6.3k & 26.8k & 116.62 $\pm$ 0.78 & 459.8 $\pm$ 3.1 & 309.1 $\pm$ 2.1 & 384.5 $\pm$ 2.6 \\
SDG zero-neg eval & 840.1M & 0.842G & 8.6k & 34.8k & 58.12 $\pm$ 0.73 & 229.6 $\pm$ 2.9 & 153.9 $\pm$ 2.0 & 191.8 $\pm$ 2.4 \\
SDG swap-sign eval & 840.1M & 0.842G & 8.6k & 34.6k & 61.47 $\pm$ 0.88 & 242.9 $\pm$ 3.5 & 162.8 $\pm$ 2.4 & 202.9 $\pm$ 3.0 \\
SDG zero-pos eval & 840.1M & 0.842G & 8.6k & 34.7k & 58.74 $\pm$ 0.69 & 232.0 $\pm$ 2.8 & 155.5 $\pm$ 1.9 & 193.8 $\pm$ 2.3 \\
Dual-Unsigned Graph & 840.1M & 0.835G & 12.1k & 40.6k & 55.46 $\pm$ 3.13 & 219.0 $\pm$ 12.4 & 146.8 $\pm$ 8.3 & 182.9 $\pm$ 10.4 \\
Unsigned Graph & 839.4M & 0.859G & 10.7k & 37.5k & 53.81 $\pm$ 1.34 & 212.7 $\pm$ 5.3 & 142.4 $\pm$ 3.6 & 177.6 $\pm$ 4.5 \\
SDG-MoE & 840.1M & 0.842G & 8.6k & 34.5k & \textbf{48.03 $\pm$ 1.81} & \textbf{189.7 $\pm$ 7.2} & \textbf{126.9 $\pm$ 4.8} & \textbf{158.3 $\pm$ 6.0} \\
\bottomrule
\end{tabular}
}
\end{table}

\begin{figure*}[h]
\centering
\includegraphics[width=\linewidth]{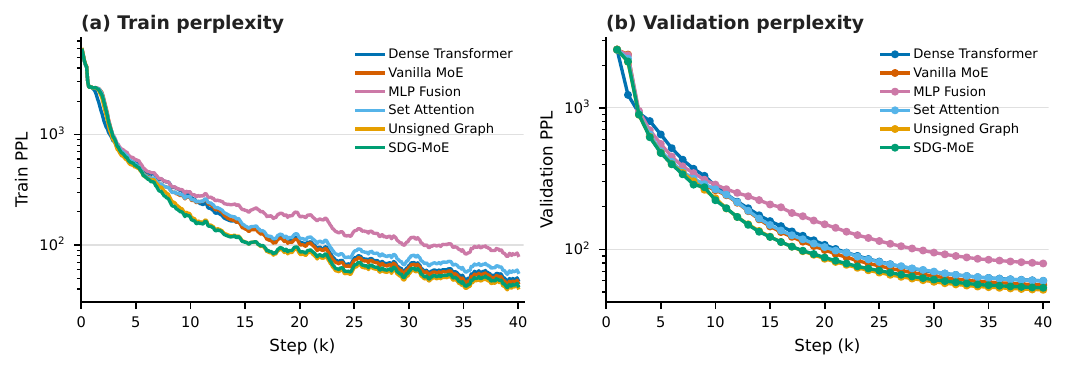}
\caption{Training and validation perplexity trajectories for the main systems. Both panels use a log-scale perplexity axis; panel (a) is smoothed training PPL and panel (b) is validation PPL at evaluation checkpoints.}
\label{fig:loss-curves}
\end{figure*}

Figure~\ref{fig:loss-curves} shows that the result is not a final-checkpoint artifact. By late training, systems are near the flat part of their validation trajectories, the relative ordering is already stable, and SDG-MoE separates from weaker baselines before the plateau.

\subsection{Ablation Study}
The SDG-specific rows in Table~\ref{tab:main-results} test whether the gain comes from signed deliberation or extra capacity. Since the learned $\lambda$ and update scale are stable on average, a natural control is to replace adaptive $\lambda g$ by a constant. This sharply worsens perplexity, showing that SDG does not merely need the right mean social strength; it needs the gate to decide which tokens, layers, and expert sets should deliberate. Evaluation-only interventions make the signed semantics more explicit. Removing critique tests whether the negative channel is redundant, removing support tests whether critique alone can carry the update, and swap-sign tests whether the labels are arbitrary after training. All three degrade the checkpoint, indicating that support and critique are not interchangeable streams. The learned signed orientation carries semantic information.

A dual-unsigned graph controls for channel count by preserving two message streams without support/critique semantics, while the unsigned graph removes signed contrast altogether. Both are worse than SDG-MoE, so the effect is not explained by parameter count, an extra graph pass, or a generic two-channel message mixer. The ablations instead support the intended mechanism: positive messages propagate compatible evidence, negative messages act as corrective contrast, and the gate decides when that correction should be active.

\begin{figure*}[h]
\centering
\includegraphics[width=\linewidth]{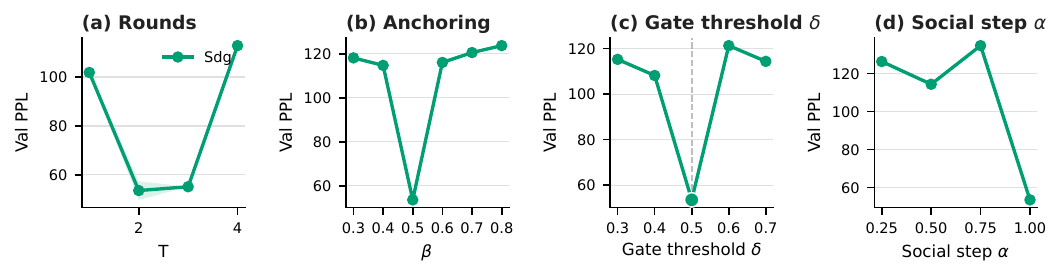}
\caption{Dynamics controls. SDG-MoE is sensitive to deliberation depth, anchoring, and social update strength, while cost grows approximately linearly with the number of rounds.}
\label{fig:dynamics-controls}
\end{figure*}

Figure~\ref{fig:dynamics-controls} studies the theory-facing hyperparameters. The best setting uses $T=2$: $T=1$ under-deliberates, while $T=4$ is worse despite more computation. This agrees with Theorem~\ref{thm:stability}: cost grows linearly in $T$, while movement toward the anchored fixed point decays geometrically, so excessive rounds can over-iterate a saturated correction. The anchoring and social-step sweeps show the same calibrated regime. Too little anchoring or too strong a social step lets the update drift; too much anchoring or an overly conservative gate suppresses useful exchange. In other words, SDG behaves like a short post-routing correction, not a deep consensus module.

\begin{figure}[h]
\centering
\includegraphics[width=\linewidth]{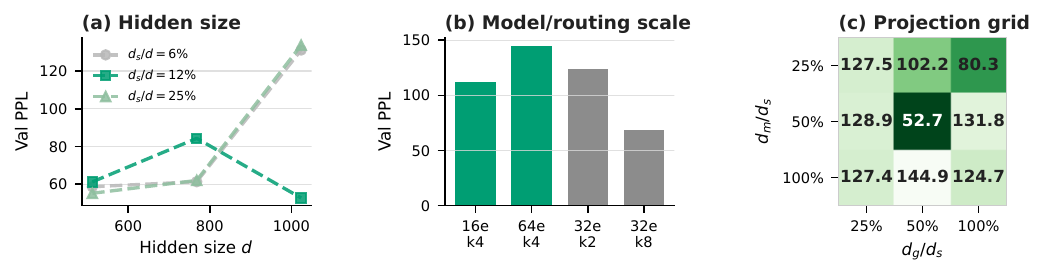}
\caption{Representation-size and scale sweeps: hidden size/shared fraction, routing-scale stress tests, and projection-ratio grid.}
\label{fig:size-scale}
\end{figure}

Figure~\ref{fig:size-scale} probes scale and bottlenecks. Panel (a) shows that hidden size alone is not monotone; at larger width, $d_s/d$ matters, and the calibrated middle fraction is best at $d=1024$. This may mean that larger representations need more data or a larger effective batch before the social block can use them reliably. Panel (b) shows routing changes are fragile without retuning: larger expert pools or smaller active sets hurt, while Top-$8$ helps only with higher active compute. Panel (c) shows that the balanced projection cell is far better than most unbalanced cells, suggesting graph and message widths must be co-sized rather than maximized independently.
\subsection{Deliberation Diagnostics}

\noindent
\begin{minipage}[t]{0.55\linewidth}
The diagnostics show active but bounded deliberation. Mean disagreement $D=0.633$ indicates that routed experts keep nontrivial diversity, while $\lambda=0.151$ makes communication selective rather than always-on. The update ratio $0.642$ shows that the social step can move the shared state, but $r_{\mathrm{shr}}=0.439$ confirms that private expert signal remains important. The graph statistics match the signed design: support has higher entropy than critique, so positive influence is broader, whereas critique is sharper and more targeted. Low ambivalence and small seed deviations suggest stable signed specialization.
\end{minipage}\hfill
\begin{minipage}[t]{0.4\linewidth}
\vspace{-1.2em}
\begin{center}
\small
\textbf{SDG deliberation metrics}\vspace{0.5em}
\setlength{\tabcolsep}{3pt}
\renewcommand{\arraystretch}{1.08}

\resizebox{0.98\linewidth}{!}{%
\begin{tabular}{@{}lccc@{}}
\toprule
Metric & Value & Std. & Meaning \\
\midrule
$D_{\mathrm{mean}}$ & 0.633 & 0.009 & expert disagreement \\
$\lambda_{\mathrm{mean}}$ & 0.151 & 0.008 & gate strength \\
$\|\Delta\|/\|h^{(0)}\|$ & 0.642 & 0.030 & relative update \\
$H(A^+)$ & 1.214 & 0.012 & support entropy \\
$H(A^-)$ & 0.646 & 0.003 & critique entropy \\
Ambiv. & 0.173 & 0.005 & sign overlap \\
$r_{\mathrm{shr}}$ & 0.439 & 0.004 & shared contribution \\
\bottomrule
\end{tabular}%
}

\vspace{0.2em}
\parbox{0.98\linewidth}{\footnotesize\centering Values are mean $\pm$ std. over 3 E0 SDG runs.}
\end{center}

\end{minipage}


\section{Conclusions}
We introduced SDG-MoE, a sparse expert layer that unifies sparse MoE computation with ideas from signed social dynamics and multi-agent consensus by turning the routed Top-$K$ set into a deliberative group before aggregation. We show theoretically that the resulting dynamics remain bounded and stable without adding significant computational overhead, and empirically that performance depends not simply on adding more communication, but on how support, critique, anchoring, bottleneck width, and deliberation depth are balanced. This opens a broader design space for post-routing expert interaction. We discuss limitations and future directions in Appendix~\ref{app:feature-work}.

\bibliographystyle{plainnat}
\bibliography{references}

\section{Limitations \& Discussions and Future Work}
\label{app:feature-work}
\paragraph{Scale and experimental scope.}
Our work is intended as a controlled study of signed expert deliberation in sparse MoE layers,
and its experimental scope is therefore deliberately focused. Experiments are conducted with relatively small models and a limited pretraining budget, which allows us to isolate the effect of the proposed communication mechanism and compare it against matched baselines under controlled conditions. While this setting provides useful evidence for the behavior of SDG-MoE, larger-scale evaluation remains an important next step. Although we investigate scaling trends for the proposed layer, it is still an open question how the same deliberation dynamics will behave in substantially larger architectures, longer training runs, and more diverse data regimes. In particular, the interaction between deliberation, routing, expert specialization, and depth may become richer as the number of layers, experts, active experts, and training tokens increases.

\paragraph{Integration with pretrained models.}
A natural next step is to evaluate SDG-MoE in the context of modern pretrained models. Rather
than training only small models from scratch, future work could insert the deliberation mechanism inside the MoE layers of an existing strong architecture and fine-tune it end-to-end. This would test whether signed expert communication remains useful when experts already encode rich pretrained representations, and whether deliberation can improve downstream performance without requiring full-scale pretraining from scratch.

\paragraph{Communication rounds and expert specialization.}
Our experiments also suggest that increasing the number of communication rounds between experts
does not lead to monotonic improvements and can sometimes degrade performance. This points to an
important design question: the most useful interaction pattern for MoE layers may not be a
deterministic procedure that repeatedly pushes all active experts toward stronger agreement. In
particular, experts do not necessarily need to be guided toward consensus at every step. Excessive
or overly regular communication may reduce specialization and turn distinct expert opinions into a more averaged signal, thereby weakening one of the main benefits of sparse expert diversity.

\paragraph{Probabilistic and stochastic deliberation.}
Instead of learning the interaction matrices as fixed deterministic structures, future work could
learn the parameters of a distribution from which such matrices are sampled. In this formulation,
edges between experts become probabilistic: the model learns not only the strength of an influence, but also the probability that a given expert participates in the interaction for a particular token. This would move SDG-MoE away from a rigid scheme in which all active experts communicate at every round, toward a stochastic deliberation process with token-dependent interaction patterns. Intuitively, such a mechanism is closer to real deliberation. Not every participant must always respond, support, or criticize the others. Sometimes an expert may remain silent, sometimes it may influence only a subset of the group, and sometimes information may travel through a short random trajectory over the expert graph. This could allow intermediate degrees of solidarity and
disagreement, rather than forcing the system toward repeated consensus updates. Architecturally, this direction resembles a random walk or stochastic debate process over the graph of active experts, where the model learns probabilistic rules of interaction instead of a single fixed communication matrix.

\section{Additional Experimental Details}
\label{app:experiment-details}
This appendix contains implementation-level settings that are held fixed across the controlled comparisons in Section~\ref{sec:experiments}, together with baseline definitions, diagnostic trajectories, additional sweeps, and proof details.

\subsection{Scaled Qwen-like Backbone}
\label{app:model-config}
When designing the experimental model, we aimed to preserve the key architectural principles of recent small-scale MoE models that achieve strong performance with a limited number of active parameters. We used Qwen3-30B-A3B \cite{qwen3} as the main reference architecture, as it combines an efficient attention configuration with sparse Mixture-of-Experts layers and represents a strong baseline among compact MoE models.

To obtain a smaller model, we scaled the main architectural components of the reference configuration by a common factor. In particular, we proportionally reduced the total number of experts, the input and output projection dimensions, and the internal dimensions of the MoE blocks. This allowed us to retain the relative structure of the original architecture while substantially reducing the total parameter count and computational cost.

At the same time, we deliberately simplified several architectural components to make the experimental setup more transparent and easier to analyze. Specifically, we removed complex variants of positional encoding, additional attention modifications, and other engineering refinements that are often used in large-scale production models. This design choice reduces the number of interacting factors and makes it easier to attribute observed effects to the core attention and MoE configuration.

The resulting architectural hyperparameters are summarized in Table~\ref{tab:model_hyperparameters}.

\begin{table}[t]
\centering
\caption{Architectural hyperparameters of the experimental MoE model.}
\label{tab:model_hyperparameters}
\begin{tabular}{p{0.34\linewidth}p{0.56\linewidth}}
\toprule
\textbf{Hyperparameter} & \textbf{Value} \\
\midrule
Number of layers & 28 \\
Hidden size & 1024 \\
Intermediate size & 9216 \\
Number of attention heads & 16 \\
Number of key-value heads & 16 \\
Attention head dimension & 64 \\
Vocabulary size & 151936 \\
Maximum sequence length & 4096 model context, 512-token training blocks \\
Position encoding & Learned \\
Attention modifications & None \\
Number of experts & 32 in the final comparisons; 16 in the routing-scale control \\
Number of active experts per token & Top-$4$ by default \\
Expert intermediate size & 288 in the final comparisons \\
Router type & Top-$k$ routing \\
Shared experts & none \\
Activation function & SiLU \\
Normalization & LayerNorm \\
Total parameters & reported by method in Table~\ref{tab:parameter-counts} \\
Active parameters & reported by method in Table~\ref{tab:parameter-counts} \\
\bottomrule
\end{tabular}
\end{table}

\begin{table}[t]
\centering
\caption{Total and active parameter counts for the compared systems. Active parameters count the parameters used on a routed forward pass under the corresponding method.}
\label{tab:parameter-counts}
\begin{tabular}{lcc}
\toprule
Method & Total parameters & Active parameters \\
\midrule
Dense Transformer & 806.22M & 806.22M \\
Vanilla MoE & 808.02M & 344.57M \\
MLP Fusion & 896.16M & 432.71M \\
Set Attention & 839.23M & 375.78M \\
Unsigned Graph & 839.45M & 376.00M \\
Dual-Unsigned Graph & 840.19M & 376.74M \\
SDG-MoE & 840.19M & 376.74M \\
GRAPHMOE & 1085.10M & 588.76M \\
Chain-of-Experts & 1073.44M & 477.82M \\
\bottomrule
\end{tabular}
\end{table}

\subsection{Baseline and Control Details}
\label{app:baseline-details}
This section specifies the implemented aggregation rules for the non-SDG
baselines and the exact interventions used in the SDG controls. For Vanilla MoE,
the layer output is the standard router-weighted expert average
\[
    y_{\mathrm{moe}}=\sum_{i\in S}w_io_i .
\]
MLP Fusion keeps the same active experts but replaces this direct average by a
small fusion network over first- and second-order expert statistics:
\[
    \mu=\sum_{i\in S}w_io_i,\qquad
    \sigma=\left(\sum_{i\in S}w_i(o_i-\mu)^2\right)^{1/2},
    \qquad
    y=\operatorname{MLP}([\mu;\sigma]).
\]
Set Attention instead treats the active expert outputs as an unordered set.
Writing $O=[o_i]_{i\in S}\in\mathbb{R}^{K\times d}$, it computes
\[
    \widetilde O=\operatorname{MHA}(O,O,O),\qquad
    y=\sum_{i\in S}w_i\widetilde o_i .
\]
Both MLP Fusion and Set Attention therefore change only the post-expert fusion
rule, whereas the router and active expert budget remain fixed.

The SDG controls are implemented as minimal modifications of the full model.
Unsigned Graph uses a single message-passing channel in place of $(A^+,A^-)$,
testing whether learned graph communication alone is sufficient. Dual-Unsigned
keeps two message channels but removes their support/critique interpretation,
testing whether gains come from signed semantics rather than simply from a wider
communication bottleneck. SDG fixed $\lambda g$ sets the effective update
coefficient to a constant value, testing whether adaptive disagreement-dependent
gating is useful beyond a fixed social step. SDG zero-neg is applied at
evaluation by setting the negative message contribution to zero in the trained
checkpoint, testing whether the learned critique channel is functionally used.

\begin{table}[h]
\centering
\caption{Default experimental hyperparameters. Ablations modify only the field under study while keeping the remaining settings fixed.}
\label{tab:experimental-hparams}
\begin{tabular}{p{0.25\linewidth}p{0.68\linewidth}}
\toprule
Group & Setting \\
\midrule
Backbone &
Qwen-like decoder-only Transformer, $d=1024$, $28$ layers, $16$ heads \\
Context and tokenizer &
Training context $512$, tokenizer checkpoint \texttt{Qwen/Qwen3-0.6B}, tied input/output embeddings \\
Data &
SmolLM-Corpus pretraining mixture with a held-out validation cache \\
Optimization &
AdamW, cosine decay, learning rate $2.5{\times}10^{-4}$, warmup $1500$ steps \\
Training budget &
$40$k optimizer steps, batch size $16$, gradient accumulation $1$ \\
Precision &
Automatic mixed precision enabled with bfloat16 \\
Sparse routing &
$32$ experts, Top-$4$ routing, expert hidden width $288$ \\
Auxiliary objective &
Router load-balancing coefficient $0.1$; signed-overlap coefficient $0$ in E0 \\
SDG communication &
$d_s=128$, graph width $d_g=64$, message width $d_m=64$, update width $128$, expert-id width $d_e=16$, disagreement width $32$ \\
SDG deliberation &
$T=2$, $\alpha=1.0$, $\beta=0.5$, $\gamma=1.0$, $m_-=2$, graph recomputed each round \\
Gate &
Disagreement-thresholded tanh gate with threshold $\delta=0.5$, initial sharpness $a=1.0$, and $\lambda_{\min}=0$ \\
Regularization ablations &
Edge dropout, support self-loop mass, and graph-input stop-gradient are off by default \\
Evaluation &
Validation perplexity, forward FLOPs/token, training and inference throughput, and external held-out perplexity \\
\bottomrule
\end{tabular}
\end{table}

\paragraph{Compute and seeds.}
Final controlled pretraining runs were executed on a single compute node with
one NVIDIA RTX6000 GPU. Unless otherwise noted, the main comparisons were run
with three independent random seeds; reported validation perplexities aggregate
these three runs.

\begin{table}[h]
\centering
\caption{Run protocol summary for the controlled comparisons.}
\label{tab:final-run-checklist}
\begin{tabular}{p{0.28\linewidth}p{0.64\linewidth}}
\toprule
Item & Setting \\
\midrule
Hardware & Single compute node with one NVIDIA RTX6000 GPU \\
Seeds & Main comparisons run with three independent random seeds \\
Training budget & $40$k optimizer steps for final runs; shorter runs for screening ablations \\
Training data & SmolLM-Corpus pretraining mixture with a held-out validation cache \\
External PPL & WikiText-103, C4 validation slice, and compact Paloma subset \\
Checkpoints & Zero-neg, swap-sign, and zero-pos are evaluation-time interventions on the trained SDG-MoE checkpoint \\
\bottomrule
\end{tabular}
\end{table}

\section{Training and Diagnostic Trajectories}
\label{app:additional-diagnostics}
\begin{figure}[h]
\centering
\includegraphics[width=\linewidth]{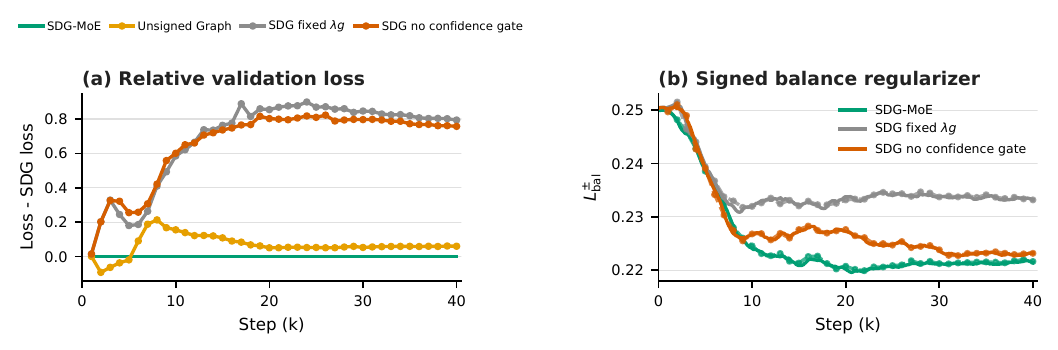}
\caption{Control loss diagnostics. Panel (a) reports validation
language-modeling loss relative to the matched SDG-MoE run; the SDG baseline is
shown as a solid zero line without markers. Panel (b) tracks the signed
support/critique balance regularizer for SDG variants; dashed curves denote
evaluation checkpoints.}
\label{fig:size-projection-plan}
\end{figure}

\begin{figure}[h]
\centering
\includegraphics[width=\linewidth]{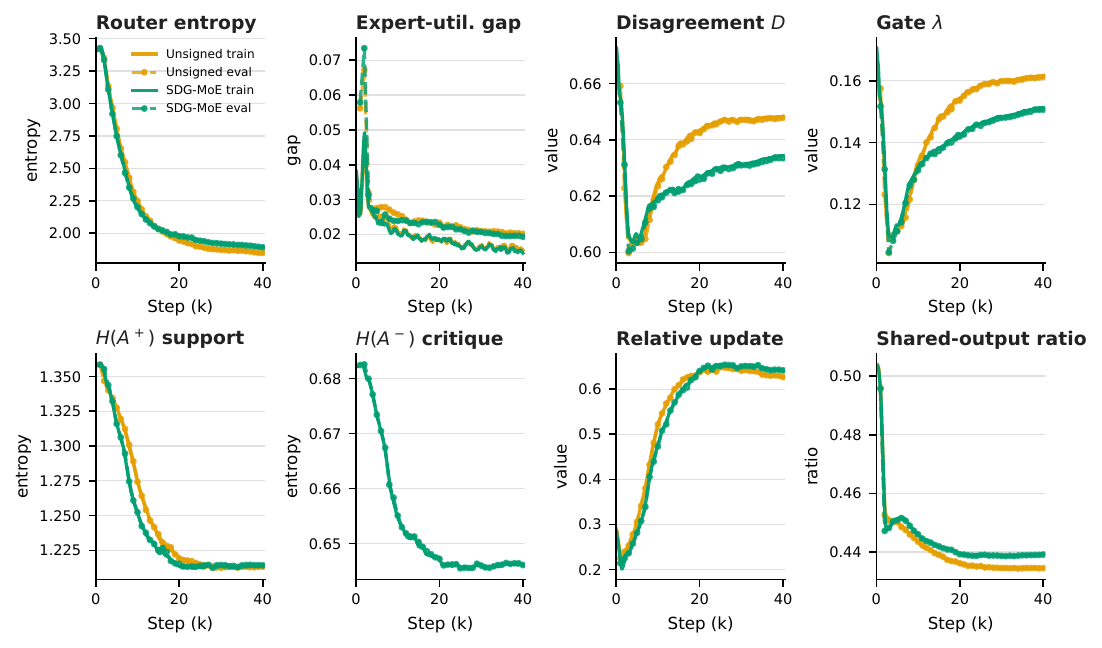}
\caption{Router and signed-deliberation diagnostics over training. Solid lines
show smoothed training metrics; dashed lines with markers show evaluation
checkpoints. The selected panels track expert utilization, gate activity,
graph entropy, update magnitude, and the shared-output contribution.}
\label{fig:dynamics-timeseries}
\end{figure}

Figure~\ref{fig:size-projection-plan} reports training losses and the signed
balance regularizer. Figure~\ref{fig:dynamics-timeseries} then tracks the
router and signed-deliberation diagnostics through training. Train curves are
lightly smoothed for readability, while evaluation checkpoints are kept raw.
These plots are meant to answer a different question from the main result
table: not only which checkpoint has lower perplexity, but whether the signed
module is active in a stable and interpretable regime while the language model
is being trained.

The relative validation-loss panel shows that the gap between SDG-MoE and its
closest controls is not a single final-checkpoint artifact. The unsigned graph
control is often close to SDG, especially later in training, but it remains
above the SDG reference for most checkpoints in this run. The dual-unsigned and
zero-negative controls are consistently worse after the early transient,
suggesting that merely adding a second communication channel is not enough and
that the learned negative channel is functionally used. The fixed-coefficient
control is the least stable among these variants, which supports the design
choice of using an adaptive disagreement-dependent social coefficient.

The signed balance regularizer is shown as a diagnostic even when its
coefficient is not the main driver of the reported result. Its trajectory is
useful because it tracks whether support and critique collapse onto the same
edges. In the current runs, the SDG variants reduce this overlap during the
first several thousand steps and then settle into a relatively narrow band.
This is consistent with the intended behavior: the negative channel becomes
structured rather than remaining a diffuse copy of the support graph. We treat
this evidence as mechanistic rather than causal, since the strongest final
claims still rely on matched controls such as zero-negative and dual-unsigned.

The router diagnostics show that all compared systems become more selective as
training progresses: router entropy decreases and the utilization gap contracts
after the early routing transient. This matters because the gains should not be
explained by a pathological router collapse; the active-expert distribution
remains broadly comparable between SDG-MoE and the unsigned graph in the
calibrated setting. The evaluation checkpoints follow the same trend as the
smoothed training curves, which suggests that the diagnostics are not an
artifact of noisy training minibatches.

The signed-deliberation panels provide a more direct picture of the social
module. Disagreement and gate activity drop sharply early in training and then
slowly rise, indicating that the model first learns coarse expert routing and
only later uses deliberation more selectively. The support entropy decreases
more strongly than the critique entropy, which matches the design intuition:
support can become concentrated around compatible experts, while the sparse
negative graph remains a targeted corrective channel. The relative update grows
over training but remains bounded, and the shared-output ratio stabilizes rather
than drifting upward without limit. Together, these trends support the
anchoring argument in the theory section: communication is present, but it does
not take over the private expert signal.

\section{Additional Sweeps}
\label{app:additional-sweeps}
\begin{figure}[h]
\centering
\includegraphics[width=\linewidth]{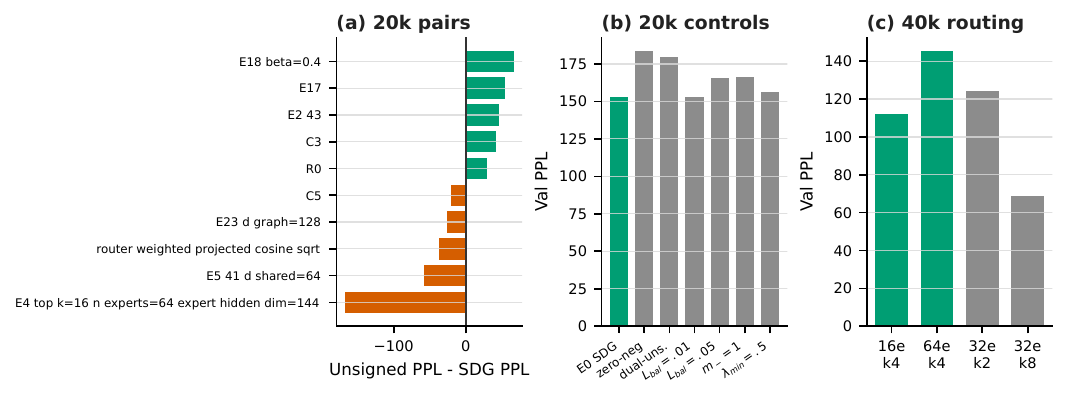}
\caption{Appendix-only sweeps. Panel (a) shows 20k paired screening deltas, where positive values favor SDG over unsigned. Panel (b) reports 20k mechanism and regularizer controls. Panel (c) shows 40k routing stress tests from the final experiment set.}
\label{fig:appendix-negative-sweeps}
\end{figure}

Figure~\ref{fig:appendix-negative-sweeps} summarizes sweeps that are informative but not central enough for the main page budget. We separate 20k screening runs from the 40k evidence used in the main text: the former are useful for seeing which mechanisms are fragile, while the latter support the final comparison. The appendix figure therefore reports paired 20k signed-vs-unsigned deltas, 20k mechanism controls, and the 40k routing stress tests.

The appendix sweeps help delimit the regime in which signed deliberation is
useful. The 20k paired runs show that signed communication is not uniformly
beneficial: it helps in some matched settings and hurts in others, which is why
the final model uses a calibrated gate rather than a stronger always-on social
update. The mechanism-control panel is similarly cautionary. Removing or
scrambling signed structure at evaluation time, adding a second unsigned
channel, or changing the signed-balance regularizer can substantially worsen
perplexity in the early screening regime.

The 40k routing stress tests are intentionally separated from the main matched
Top-$4$ comparison because they change the active-expert budget or expert pool.
They are still informative as stress tests: increasing the number of active
experts can improve perplexity, but it also changes the compute profile enough
that it should not be used to claim a matched-budget gain. These mixed outcomes
are why the main paper emphasizes a short, bounded post-routing correction with
adaptive gating rather than a uniformly stronger communication block.

Overall, the additional sweeps support a conservative interpretation. SDG-MoE
does not improve merely because the social block is made stronger, deeper, or
more heavily regularized. The useful configuration is a short, bounded
post-routing correction with enough shared capacity to communicate, a sparse
critique channel, and an adaptive gate that avoids applying the social update
uniformly to every token.

\section{Future Large-Scale Evaluation}
\label{app:planned-eval}
The current experiments are deliberately controlled and small enough to permit
many matched ablations. Larger-scale evaluation should move from broad
screening toward a smaller matrix of longer runs: Vanilla MoE, Unsigned Graph,
SDG-MoE, and the strongest interaction baselines under the same effective batch
and hardware. External held-out perplexity should remain the primary transfer
metric at this scale, with downstream reasoning benchmarks reported as
secondary sanity checks unless the pretraining budget is large enough to make
those scores stable.

\section{One-Round SDG-MoE Deliberation}
\label{app:algorithm}
Algorithm~\ref{alg:sdg-round} expands the compact update in Section~\ref{sec:architecture} into the operations used by one deliberation round. We write it for a single token after the router has selected its active expert set $S$; in the implementation the same computation is batched over tokens and layers. The algorithm emphasizes the three points that are most relevant for reproducibility: positive and negative graphs are constructed from the current shared states, the global disagreement gate modulates the social update, and anchoring mixes the updated state with the initial shared representation. Multi-round deliberation simply repeats this procedure, optionally rebuilding the signed graph after each round.

\begin{algorithm}[t]
\caption{One round of signed deliberation for one token}
\label{alg:sdg-round}
\begin{algorithmic}[1]
\Require Shared states $H^{(t)}=\{h_{i,\mathrm{shr}}^{(t)}\}_{i\in S}$, initial states $H^{(0)}$, expert ids $\{e_i\}_{i\in S}$, token state $x$, router weights $\{w_i\}_{i\in S}$
\Ensure Updated shared states $H^{(t+1)}$
\For{$i\in S$}
    \State $z_i^{(t)} \gets [\,\operatorname{LN}(h_{i,\mathrm{shr}}^{(t)}); e_i\,]$
    \State $q_i^+ \gets W_Q^+z_i^{(t)},\quad k_i^+ \gets W_K^+z_i^{(t)}$
    \State $q_i^- \gets W_Q^-z_i^{(t)},\quad k_i^- \gets W_K^-z_i^{(t)}$
    \State $\widehat h_i^{(t)} \gets P_D h_{i,\mathrm{shr}}^{(t)}/(\norm{P_D h_{i,\mathrm{shr}}^{(t)}}_2+\varepsilon)$
\EndFor
\State $A^+ \gets \softmax_j(\langle q_i^+,k_j^+\rangle/\sqrt{d_g}+B^+_{ij})$
\State $\widetilde A^- \gets \softmax_j(\langle q_i^-,k_j^-\rangle/\sqrt{d_g}+B^-_{ij})$
\State $A^- \gets \operatorname{Normalize}(\TopM_{m_-}(\widetilde A^-))$
\State $D^{(t)} \gets \sqrt{\frac{1}{K(K-1)}\sum_{i\ne j}\frac{1-\langle \widehat h_i^{(t)},\widehat h_j^{(t)}\rangle}{2}}$
\State $\lambda^{(t)} \gets \lambda_{\min}+(1-\lambda_{\min})\tanh(a[D^{(t)}-\delta]_+)$
\For{$i\in S$}
    \State $g_i(x)\gets \sigma(w_{g,i}^{\top}x+b_{g,i})$ \Comment{or $1$ when the confidence gate is disabled}
\EndFor
\For{$j\in S$}
    \State $u_j^{(t)}\gets Vh_{j,\mathrm{shr}}^{(t)}$
\EndFor
\For{$i\in S$}
    \State $m_i^+ \gets \sum_{j\in S}A_{ij}^+u_j^{(t)},\quad m_i^- \gets \sum_{j\in S}A_{ij}^-u_j^{(t)}$
    \State $\Delta_i^{(t)} \gets U_2\phi(U_1[\,h_{i,\mathrm{shr}}^{(t)};m_i^+;m_i^+-\gamma m_i^-\,]+b_1)+b_2$
    \State $\bar h_{i,\mathrm{shr}}^{(t+1)}\gets h_{i,\mathrm{shr}}^{(t)}+\alpha\lambda^{(t)}g_i(x)\Delta_i^{(t)}$
    \State $h_{i,\mathrm{shr}}^{(t+1)}\gets \beta h_{i,\mathrm{shr}}^{(0)}+(1-\beta)\bar h_{i,\mathrm{shr}}^{(t+1)}$
\EndFor
\State \Return $H^{(t+1)}$
\end{algorithmic}
\end{algorithm}

\section{Proof Details and Gate Calibration}
\label{app:theory}

\subsection{Local Assumptions}
\label{app:assumptions}
The assumptions in Section~\ref{sec:theory} match the finite-round computation used by SDG-MoE. Once the router selects an active set and the Top-$m_-$ criticism supports are fixed, one deliberation round is a composition of affine maps, LayerNorm with numerical $\varepsilon$, softmax, bounded gates, and standard activations. Hence, on the bounded region visited by a finite deliberation trajectory, we use constants
\[
    \norm{\Delta_x(H)}_{\mathrm F}\le B_\Delta,\quad
    \norm{\Delta_x(H)-\Delta_x(\widetilde H)}_{\mathrm F}
    \le L_\Delta\norm{H-\widetilde H}_{\mathrm F},
\]
with $D$ and $\lambda$ Lipschitz with constants $L_D$ and $L_\lambda$, $0\le g_i(x)\le g_{\max}$ for every active expert, and $\max_i\norm{W^{\mathrm{out,shr}}_i}_2\le B_W$. These local constants are sufficient because SDG-MoE uses only a small fixed number of rounds. Exact routing or Top-$m_-$ ties have measure zero under continuous logits, so the estimates can be read conditionally on a fixed active support. If a global bounded-update certificate is desired, the update field can be clipped as
\[
    \Delta_x(H)\leftarrow
    B_\Delta \Delta_x(H)/\max\{B_\Delta,\norm{\Delta_x(H)}_{\mathrm F}\}.
\]
This keeps the finite-round interpretation unchanged while making boundedness explicit.

\subsection{Proof of Proposition~\ref{prop:cost}}
\label{app:cost-proof}
One SDG round consists of graph construction, disagreement estimation, message aggregation, and the update MLP. Positive and negative query/key projections cost $O(K(d_s+d_e)d_g)$ and their $K\times K$ dot products cost $O(K^2d_g)$. Disagreement uses $K$ projections into $d_{\mathrm{dis}}$ and pairwise cosine scores, giving $O(Kd_sd_{\mathrm{dis}}+K^2d_{\mathrm{dis}})$. Message projection and aggregation cost $O(Kd_sd_m+K^2d_m)$. The update MLP maps each vector of dimension $d_s+2d_m$ into width $d_{\mathrm{upd}}$ and back to $d_s$, costing $O(K(d_s+2d_m)d_{\mathrm{upd}}+Kd_{\mathrm{upd}}d_s)$. Collecting widths into
\[
    b=\max\{d_s,d_g,d_m,d_{\mathrm{dis}},d_{\mathrm{upd}},d_e\}
\]
gives $O(Kb^2+K^2b)$ per round and $O(T(Kb^2+K^2b))$ for $T$ rounds. With shared round parameters, the parameter count contains the graph projections, disagreement projection, message projection, update MLP, and the expert-id table for the $N$ experts in the layer, hence $O(b^2+Nd_e)$. Dividing by the active expert FFN cost $\Theta(Kdd_{\mathrm{ff}})$ proves Eq.~\eqref{eq:relative-cost}.

\subsection{Proof of Theorem~\ref{thm:stability}}
\label{app:stability-proof}
Let $R_t=\norm{H^{(t)}-H^{(0)}}_{\mathrm F}$. Subtracting $H^{(0)}$ from Eq.~\eqref{eq:abstract-sdg} gives
\[
    H^{(t+1)}-H^{(0)}
    =
    (1-\beta)(H^{(t)}-H^{(0)})
    +(1-\beta)\alpha\lambda(D(H^{(t)}))G(x)\Delta_x(H^{(t)}).
\]
Using $0\le\lambda\le1$, $\norm{G(x)}_2\le g_{\max}$, and $\norm{\Delta_x(H)}_{\mathrm F}\le B_\Delta$,
\[
    R_{t+1}\le (1-\beta)R_t+(1-\beta)\alpha g_{\max}B_\Delta .
\]
Since $R_0=0$, unrolling the scalar recursion yields Eq.~\eqref{eq:bounded-hidden}. The output bound follows from
\[
    y_T-y_0=\sum_{i\in S}w_i W^{\mathrm{out,shr}}_i
    \left(h_{i,\mathrm{shr}}^{(T)}-h_{i,\mathrm{shr}}^{(0)}\right),
\]
the nonnegativity and unit sum of router weights, and $\max_i\norm{W^{\mathrm{out,shr}}_i}_2\le B_W$.

For contraction, define
\[
    F(H)=\beta H^{(0)}
    +(1-\beta)[H+\alpha\lambda(D(H))G(x)\Delta_x(H)] .
\]
For two states $H,\widetilde H$ in the operating region,
\[
\begin{aligned}
    \norm{F(H)-F(\widetilde H)}_{\mathrm F}
    &\le (1-\beta)\norm{H-\widetilde H}_{\mathrm F}\\
    &\quad +(1-\beta)\alpha g_{\max}
    \norm{\lambda(D(H))\Delta_x(H)
    -\lambda(D(\widetilde H))\Delta_x(\widetilde H)}_{\mathrm F}.
\end{aligned}
\]
The second norm is bounded by
\[
    (L_\Delta+B_\Delta L_\lambda L_D)\norm{H-\widetilde H}_{\mathrm F}=C\norm{H-\widetilde H}_{\mathrm F},
\]
which gives the Lipschitz constant $q=(1-\beta)(1+\alpha g_{\max}C)$. If $q<1$, Banach's fixed-point theorem gives a unique fixed point and linear convergence. Written as design constraints, the same condition is
\[
    \alpha g_{\max}C < \frac{\beta}{1-\beta}
    \qquad\Longleftrightarrow\qquad
    \beta > \frac{\alpha g_{\max}C}{1+\alpha g_{\max}C},
\]
or, for fixed $\beta$ and $C$,
\[
    \alpha < \frac{\beta}{(1-\beta)g_{\max}C}.
\]
Thus stronger social steps require either stronger anchoring or a smaller Lipschitz constant for the deliberation field.

\subsection{Proof of Proposition~\ref{prop:targeted-critique}}
\label{app:signed-proof}
Work in coordinates translated so that the inlier center is the origin. Thus the
$K-1$ inlier messages satisfy $u_j=\epsilon_j$ with
$\norm{\epsilon_j}\le\epsilon$, and the outlier is $u_o=r$. This centering
removes the otherwise uncancelled translation term
$(\sum_j A^+_{ij}-\gamma\sum_j A^-_{ij})c=(1-\gamma)c$ that would appear before
translation. Let $c_i=m_i^+-\gamma m_i^-$ be the signed contrast. If an inlier
$i$ assigns $A_{io}^-\ge\eta$ and $A_{io}^+\le\xi$, then the outlier contributes
at most $(\xi-\gamma\eta)r$ to $c_i$. All remaining centered inlier terms have
projection on $r$ bounded by $(1+\gamma)\epsilon\norm r$ because positive and
negative rows each have total mass at most one. Therefore
\[
    \langle c_i,r\rangle
    \le
    (1+\gamma)\epsilon\norm r
    -(\gamma\eta-\xi)\norm r^2 .
\]
When $\gamma\eta>\xi$ and $\norm r>(1+\gamma)\epsilon/(\gamma\eta-\xi)$, the projection is negative, so the signed contrast points away from the outlier direction.

\subsection{Gate Calibration}
\label{app:gate-calibration}
For unit projected states, Eq.~\eqref{eq:disagreement} gives
\[
    D^2
    =
    \frac{1}{K(K-1)}
    \sum_{i\ne j}\frac{1-\langle \widehat h_i,\widehat h_j\rangle}{2}
    =
    \frac12(1-\overline{\cos}).
\]
Since the sum of all pairwise squared distances among $K$ unit vectors is bounded, $0\le D\le \sqrt{K/(2(K-1))}$, and in practice the score is clipped by the cosine-distance construction. For the default gate
\[
    \lambda(D)=
    \lambda_{\min}+(1-\lambda_{\min})\tanh(a[D-\delta]_+),
\]
the Lipschitz constant is at most $(1-\lambda_{\min})a$. Combining this with Theorem~\ref{thm:stability} gives the conservative stable-gate condition
\[
    (1-\beta)\left[
    1+\alpha g_{\max}
    \left(L_\Delta+B_\Delta(1-\lambda_{\min})aL_D\right)
    \right]<1 .
\]
Equivalently, for fixed $\alpha$ and $\beta$, the gate sharpness should satisfy
\[
    a <
    \frac{
    \frac{\beta}{(1-\beta)\alpha g_{\max}} - L_\Delta
    }{
    B_\Delta(1-\lambda_{\min})L_D
    },
\]
whenever the numerator is positive. This motivates tuning $\alpha$, $\beta$, and gate sharpness $a$ jointly rather than independently: sharper gates make $\lambda(D)$ more sensitive to small disagreement changes and therefore require either smaller social steps or stronger anchoring for the same conservative stability certificate.

\newpage
\end{document}